\documentclass[fleqn,10pt]{wlscirep}
\usepackage[utf8]{inputenc}
\usepackage[T1]{fontenc}
\usepackage{enumitem}
\usepackage{setspace}
\usepackage{multicol}
\usepackage{float}
\usepackage{booktabs}
\usepackage{array}
\usepackage{hyperref}
\usepackage{cleveref}
\usepackage{amsmath}
\usepackage{caption}
\usepackage{fancyhdr}
\usepackage{multirow}
\title{AVPDN: Learning Motion-Robust and Scale-Adaptive Representations for Video-Based Polyp Detection}

\author[1,*]{Zilin Chen}
\author{Shengnan Lu}

\affil{School of Computer Science, Xi'an Shiyou University, Xian, 710065, China}

\affil[*]{chenzilin0925@163.com}

\begin{abstract}
Accurate detection of polyps is of critical importance for the early and intermediate stages of colorectal cancer diagnosis. Compared to static images, dynamic colonoscopy videos provide more comprehensive visual information, which can facilitate the development of effective treatment plans. However, unlike fixed-camera recordings, colonoscopy videos often exhibit rapid camera movement, introducing substantial background noise that disrupts the structural integrity of the scene and increases the risk of false positives.
To address these challenges, we propose the Adaptive Video Polyp Detection Network (AVPDN), a robust framework for multi-scale polyp detection in colonoscopy videos. AVPDN incorporates two key components: the Adaptive Feature Interaction and Augmentation (AFIA) module and the Scale-Aware Context Integration (SACI) module.
The AFIA module adopts a triple-branch architecture to enhance feature representation. It employs dense self-attention for global context modeling, sparse self-attention to mitigate the influence of low query-key similarity in feature aggregation, and channel shuffle operations to facilitate inter-branch information exchange.
In parallel, the SACI module is designed to strengthen multi-scale feature integration. It utilizes dilated convolutions with varying receptive fields to capture contextual information at multiple spatial scales, thereby improving the model’s denoising capability.
Experiments conducted on several challenging public benchmarks demonstrate the effectiveness and generalization ability of the proposed method, achieving competitive performance in video-based polyp detection tasks.

\end{abstract}

\begin{document}

\flushbottom
\maketitle
%
%
\thispagestyle{empty}

\section*{Introduction}

Colorectal cancer (CRC) represents a major global health burden, ranking as the third most commonly diagnosed cancer and the second leading cause of cancer-related mortality. The vast majority of colorectal malignancies originate from adenomatous polyps~\cite{r1}, making the accurate detection of polyps critically important for early and intermediate-stage diagnosis.
In recent years, significant efforts have been devoted to improving the understanding and interpretation of colonoscopic examinations. Traditional approaches to polyp detection rely heavily on manual visual inspection and documentation, which are inherently time-consuming, labor-intensive, and susceptible to inter-observer variability. As a result, computer-aided systems for automated polyp detection and analysis in colonoscopy have gained increasing attention in the field of medical image analysis~\cite{r27}.
These systems can assist clinicians in identifying precancerous lesions, particularly polyps, during early gastrointestinal endoscopic procedures. Moreover, they provide quantitative information such as polyp size, shape, type, and anatomical location, thereby supporting the development of personalized treatment strategies tailored to individual patient profiles.

\begin{figure}[t]
\centering
\includegraphics[width=1\textwidth,height=0.17\textwidth]{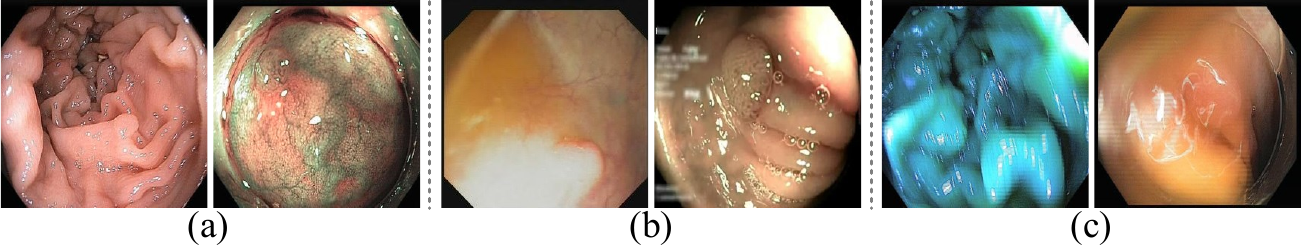}
\caption{(a) Motion Artifacts: High-speed camera movement leads to blurring, distortion, or fracture of background tissues (e.g., intestinal folds, vascular textures). (b) Transient Interference: Dynamic contact between the endoscope and intestinal tissue generates air bubbles and specular highlights. (c) Multi-scale Background Confusion: Rapid camera displacement results in significant scale variations of the same anatomical structure.}
\label{intro}
\end{figure}
Before the widespread adoption of artificial intelligence, polyp detection primarily relied on manual visual assessment. This approach was constrained by subjective interpretation, limited field of view, and dependence on the clinician's expertise. With the rapid development of deep learning, particularly Convolutional Neural Networks (CNNs)~\cite{r28} and Transformer architectures~\cite{r29}, numerous automated detection models have been proposed, significantly advancing the field. These systems aim to assist endoscopists and reduce the likelihood of misdiagnosis~\cite{r30}. Transformer-based object detection methods have emerged as a promising direction by leveraging robust semantic modeling and self-attention mechanisms. Among these, DETR (DEtection TRansformer)~\cite{r31} marks a seminal milestone, reformulating object detection as a set prediction task and eliminating the need for anchor boxes and Non-Maximum Suppression (NMS)~\cite{r32}.

Despite these advancements, accurate and reliable polyp detection remains a challenging task. Most detection models are derived from experience in natural image or video object detection and often overlook the unique characteristics of colonoscopic data, which exhibit high variability and complexity. Unlike natural scenes typically captured by static cameras, colonoscopy involves dynamic environments with continuous camera movement, posing significant challenges for automated detection.
As illustrated in Figure~\ref{intro}, several domain-specific factors hinder accurate detection. First, the rapid movement of the endoscopic device, combined with intestinal peristalsis, results in motion blur and frame discontinuities, especially in background textures such as mucosal folds and vascular structures. Second, transient visual artifacts, including specular reflections, air bubbles, and fluid, frequently appear due to contact between the endoscope and the intestinal wall, further degrading inter-frame consistency. Third, the varying distance between the camera and the target induces substantial scale changes for the same anatomical structures. Moreover, small polyps may visually resemble surrounding tissues and be treated as occluded objects, leading to frequent false positives or false negatives during inference.

To address these challenges, we propose the Adaptive Video Polyp Detection Network (AVPDN), a framework tailored for robust multi-scale polyp detection in colonoscopy videos. The proposed architecture integrates two key components: the Adaptive Feature Interaction and Augmentation (AFIA) module and the Scale-Aware Context Integration (SACI) module.
The AFIA module employs a dual-branch adaptive self-attention mechanism to capture long-range dependencies and refine feature aggregation under complex motion conditions. Additionally, it incorporates a channel shuffle operation to encourage inter-channel information exchange, enhancing feature diversity. The SACI module further strengthens the network's capability to extract multi-scale features by utilizing dilated convolutions to integrate both coarse global context and fine local details.
Experimental evaluations conducted on publicly available and challenging benchmarks demonstrate that the proposed method achieves strong detection performance and generalization capability across various scenarios.

In summary, the main contributions of this work are as follows:
\begin{itemize}
    \item We propose the Adaptive Video Polyp Detection Network (AVPDN), a framework specifically designed to address the challenges of multi-scale polyp detection in dynamic colonoscopy videos.
    \item We introduce the Adaptive Feature Enhancement (AFE) Encoder, which integrates the AFIA and SACI modules to adaptively refine feature representations and capture the spatial distribution of polyps across scales.
    \item Extensive experiments on public datasets confirm that AVPDN consistently outperforms existing methods in terms of detection accuracy and robustness.
\end{itemize}

\section*{Related Work}
In this section, we will introduce the progress of colonoscopy video polyp detection algorithms from two categories: hand-crafted and deep learning.

\subsection*{Polyp Detection Based on Hand-crafted Feature}
Real-time polyp detection has been a prominent area of research for several decades as a computer-aided diagnostic technology for clinical endoscopy. Similar to the early stages of traditional object detection, most polyp detection methods have historically involved extracting hand-crafted features on color, shape, and texture through conventional image processing techniques to identify candidate polyp regions. Tajbakhsh et al.~\cite{r2} proposed a computer-aided colonoscopy polyp detection system based on shape methods, which removed non-polyp structures and used shape information to reliably locate polyps. Sasmal et al.~\cite{r3} proposed a method for classifying colonic polyps using fused geometry, texture, and color features with SVM and RUSBoosted tree classifiers is proposed and evaluated. Ren et al.~\cite{r4} proposed to combine the shape index and the multi-scale enhancement filter by Gaussian smooth distance field to generate the candidate polyp. The limited representational power and the labor-intensive design of these methods restrict their ability to handle complex background variations and noise interference effectively.

\begin{figure}[t]
    \centering
    \includegraphics[width=1\textwidth,height=0.4\textwidth]{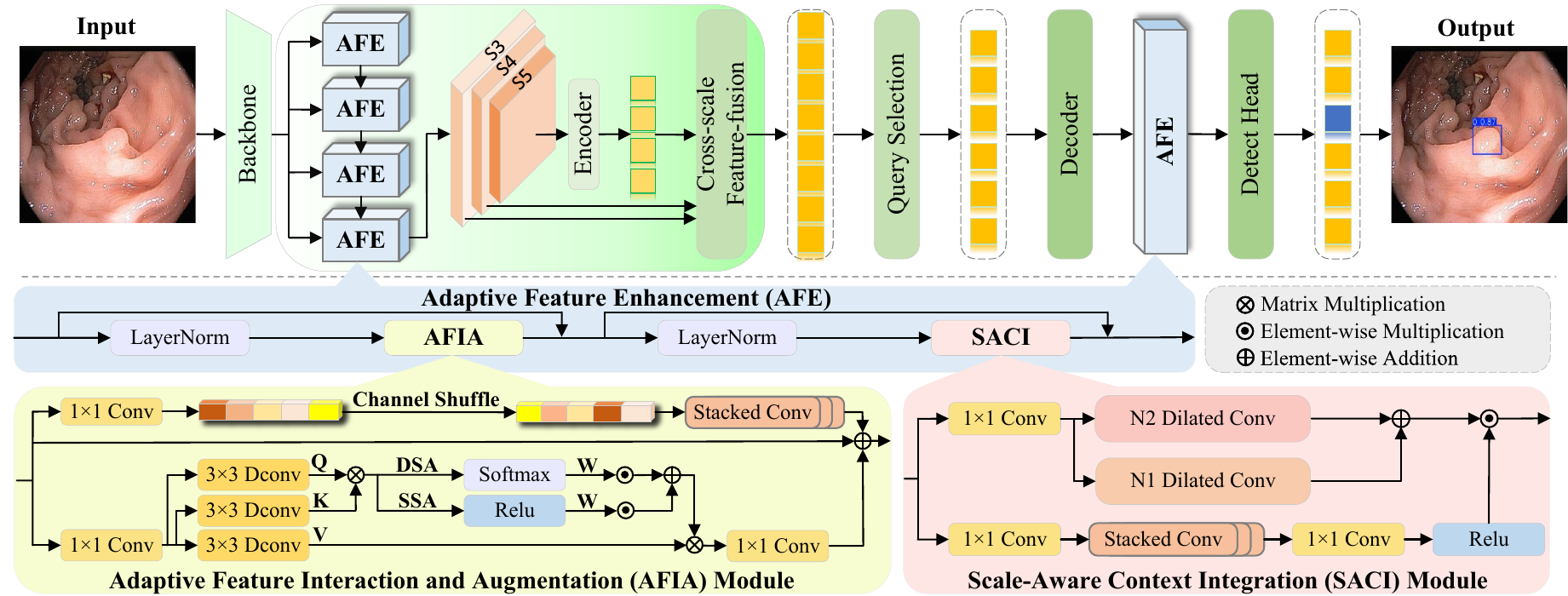}
\caption{Detailed illustration of our Adaptive Video Polyp Detection Network (AVPDN) for video polyp detection. The top portion illustrates the framework of AVPDN.The network processes input images via a backbone and a series of Adaptive Feature Enhancement (AFE) blocks. The Adaptive Feature Interaction and Augmentation (AFIA) module within each AFE block employs dense and sparse self-attention to learn contextual dependencies within feature maps, alongside channel shuffling to facilitate information flow. After this, the Scale-Aware Context Integration (SACI) module uses multiple dilated convolution layers to incorporate context at varied scales, thereby improving the model’s multi-scale understanding.}
\label{frame}
\end{figure}

\subsection*{Polyp Detection Based on Deep Learning}  
Recent achievements in deep learning have markedly accelerated the development of artificial intelligence in medicine, thereby stimulating greater interdisciplinary scholarly inquiry. Consequently, numerous polyp detection approaches have emerged and have been applied to computer-aided diagnosis. In the early stage, colonoscopy polyp detection is considered as object detection, where those methods are generally grouped into two categories: two-stage detection and one-stage detection~\cite{r5}. As a classic two-stage object detection method, Faster R-CNN’s~\cite{r6} Region Proposal Network (RPN) is a significant milestone, enabling end-to-end training of region proposal generation and feature extraction.  Existing polyp detectors generally adopt the
CNN-based architecture, the most famous of which is the
YOLO~\cite{r7},~\cite{r8},~\cite{r9},~\cite{r10} detectors due to their reasonable trade-off between speed and accuracy. Jiang et al.~\cite{r11} proposed YONA, a novel end-to-end framework, that addresses unstable training in video polyp detection by adaptively aligning features from adjacent frames and using contrastive learning. Bernal et al.~\cite{r12} proposed a triple-stage approach for automatic polyp detection in colonoscopy videos, involving region segmentation, a novel descriptor (SA-DOVA), and region classification based on maximal descriptor values. The rapid camera motion typical in colonoscopy videos introduces substantial noise, which these methods fail to appropriately compensate for, thereby potentially compromising model performance. Differently, our AVPDN aims to deal with challenging polyp samples through improving multi-scale feature aggregation and denoising efficacy.

\section*{Proposed Method}\label{sec:method} 
\subsection*{Overview}
The overall pipeline of our proposed Adaptive Video Polyp Detection Network (AVPDN) framework is meticulously illustrated in Figure \ref{frame}. We leverage the Real-Time DEtection TRansformer (RT-DETR)~\cite{r13} as the foundation of our detector, capitalizing on its inherent speed and accuracy advantages. AVPDN builds upon this architecture, incorporating several key modifications tailored for the unique challenges of polyp detection in dynamic colonoscopy videos. Specifically, AVPDN consists of a ResNet-50~\cite{r27} backbone for feature extraction, custom-designed Adaptive Feature Enhancement (AFE) blocks for robust feature representation, and a Transformer encoder-decoder (inherited from RT-DETR) with auxiliary prediction heads for accurate polyp localization and classification.

The processing begins with the current frame of each colonoscopy video clip, considered as an input image $I \in \mathbb{R}^{H \times W \times C}$, which we denote as “Input”. This input image is initially fed into the ResNet-50 backbone, pre-trained on ImageNet, to extract hierarchical feature maps. The ResNet-50 architecture is carefully chosen for its balance between representational capacity and computational efficiency. The output of the backbone is a series of low-level feature maps $F_0\in \mathbb{R}^{H/s\times W/s \times C'}$, where $s$ is the downsampling factor determined by the backbone’s architecture (typically 32), $H \times W$, $C'$ are the image resolution and the number of channels in the feature maps, respectively. These low-level features capture basic visual information such as edges, textures, and color variations, but lack the high-level contextual understanding necessary for robust polyp detection.

To address this limitation, the low-level representation $F_0$ then passes through a series of Adaptive Feature Enhancement (AFE) blocks. Each AFE block is designed to progressively refine the feature representation, mitigating noise and enhancing polyp-specific features. Importantly, these AFE blocks are applied iteratively at multiple scales within the feature pyramid to progressively enhance features across different resolutions. This multi-scale processing is crucial for detecting polyps of varying sizes, which is a common challenge in colonoscopy videos. Within each AFE Block, the input features $F_0$ (either $F_0$ for the first block or the output of the previous AFE block) first pass through Layer Normalization. Layer Normalization is applied to stabilize the input distribution, improving the training process and generalization performance. Subsequently, these normalized features are processed by the Adaptive Feature Interaction and Augmentation (AFIA) module, a core contribution of our work, designed to enhance feature discriminability and expressiveness, thereby suppressing background noise interference. Following another Layer Normalization step, multi-scale information is aggregated via the Scale-Aware Context Integration (SACI) module, another critical component. The function of SACI is to combine fine-grained features from various points. Detailed descriptions of the AFIA and SACI are provided in III.B. By applying layers of iterative transformation to features, the noise from colonoscopy videos is reduced, and key aspects are highlighted.

After the AFE Blocks, the multi-scale features are then fed into the Transformer decoder, which adopts a deformable attention mechanism~\cite{r26}. The deformable attention mechanism learns the characteristics of polyps within video frames to generate better object queries and better predictions of bounding boxes~\cite{r13}. For detection, we adopt GIoU Loss and L1 Loss. Given prediction bounding boxes $B_p$ and ground
truth bounding boxes $B_g$, the overall loss function $L$ can be formulated as:

\begin{equation}
\text{GIoU} = \frac{\text{Area}(B_p \cap B_g)}{\text{Area}(B_p \cup B_g)} - \frac{\text{Area}(C \setminus (B_p \cup B_g))}{\text{Area}(C)}
\end{equation}
\begin{equation}
L_{GIoU} = 1 - \text{GIoU}
\end{equation}
\begin{equation}
\begin{split}
L_1 = \frac{1}{n} \sum_{i=1}^{n} ( |x_p^i - x_g^i| + |y_p^i - y_g^i| + |w_p^i - w_g^i| + |h_p^i - h_g^i| )
\end{split}
\end{equation}
\begin{equation}
\begin{aligned}
    L &= L_{GIoU}(B_p,B_g)+L_1(B_p,B_g), \\
\end{aligned}
\end{equation}
where $C$ is the smallest enclosing box for both $B_p$ and $B_g$, $n$ is the number of bounding boxes (typically 1, or if averaged), $x_p^i,y_p^i,w_p^i,h_p^i$ and $x_g^i,y_g^i,w_g^i,h_g^i$ are the coordinates of the predicted bounding box $B_p$ and $B_g$ for the 
$i$-th box.

\subsection*{Adaptive Feature Enhancement (AFE) Block Design}  
\subsubsection*{Adaptive Feature Interaction and Augmentation (AFIA) Module} As previously discussed, the Adaptive Feature Interaction and Augmentation (AFIA) module lies at the heart of our proposed AFE block, as illustrated in Figure \ref{frame}. In the polyp detection task, traditional Vision Transformers (ViT)~\cite{r14} consider all tokens inside the feature map in colorectal video backgrounds, while powerful for capturing long-range dependencies, often struggle in the context of colonoscopy videos. These videos are characterized by complex textures, motion blur, specular reflections, and scale variations, leading to a high degree of noise and redundancy in the feature maps. Directly applying standard ViTs results in inefficient computation due to irrelevant regions being processed and the introduction of redundant features that degrade model performance. To overcome these limitations, the AFIA module is meticulously engineered to adaptively enhance relevant features while suppressing noise and redundancy, enabling more robust and accurate polyp detection. The AFIA module achieves this through a carefully orchestrated three-pronged approach: dual-branch adaptive self-attention, channel shuffling for feature diversification, and adaptive feature fusion for optimized information integration.

\begin{table}
    \centering
    \caption{Quantitative comparison of different image/video-based detectors with our AVPDN on LDPolypVideo dataset and CVC-VideoClinicDB database. AP means Average Precision.}
    \label{tab1}
    \begin{tabular}{l|cccc|cccc|c}
        \toprule
        \multirow{2}{*}{ Image/Video-based Detectors } & \multicolumn{4}{c|}{LDPolypVideo} & \multicolumn{4}{c|}{CVC-VideoClinicDB} &\multicolumn{1}{c}{}\\
        \cmidrule(lr){2-5} \cmidrule(lr){6-10}
                             & AP   & Precision & Recall & F1-Score  & AP   & Precision & Recall & F1-Score & FPS\\

        \midrule
        Faster R-CNN~\cite{r18}& 73.2 & 77.2      & 69.6   & 73.2     & 88.2 & 84.6      & \textbf{98.2}   & 90.9     &43.1\\
        Sparse R-CNN~\cite{r19}& 65.3 & 72.0      & 50.5   & 59.5     & 87.9 & 85.1      & 96.4   & 90.4     &44.5\\
        Center Net~\cite{r20}  & 69.7 & 74.6      & 65.4   & 69.9   & 87.0 & 92.0      & 80.5   & 85.9     &50.0\\
 FGFA~\cite{r35}& 86.9& 87.7& 85.4& 86.5& 92.0& 93.9& 91.1& 92.4&1.8\\
        YOLOv5~\cite{r21}      & 92.0 & 92.2      & 91.4   & 91.7  & 87.5 & 91.4      & 83.5   & 87.3   &53.1\\
        YOLOv8      & 92.9 & 93.1      & 91.6   & 92.3  & 90.9 & 92.2      & 87.8   & 89.9   &54.0\\
        YOLO X                 & 90.7 & 89.9      & 91.0& 90.4& 86.7 & 92.6      & 85.6   & 89.0    &52.7\\
        YOLOv11                & 91.9& 92.2      & 86.3   & 89.1  & 92.1 & 91.7      & 92.2& 91.8&54.6\\
         YOLOv12               & 92.5& 92.6& 91.6& 92.0& 93.1 & 93.9& 93.3& 93.5&53.9\\
        TransVOD~\cite{r24}& 74.1 & 79.3      & 69.6   & 74.1     & 90.5 & 92.1      & 91.4   & 91.7  &8.9\\
        DINO~\cite{r25}    & 89.6& 89.7      & 88.3   & 88.9     & 91.3 & 93.6      & 90.1   & 91.8   &39.8\\
        Deformable DETR~\cite{r26}        & 90.5 & 90.5      & 89.9   & 90.2     & 85.4 & 91.8      & 79.6   & 85.2&48.9\\
        RT-DETR~\cite{r13}  & 94.2 & 94.4& 92.3& 93.3     & 93.1 & 94.1      & 93.9   & 93.9   &52.1\\
        \midrule
        AVPDN (Ours)   & \textbf{96.6} & \textbf{96.8}& \textbf{95.0}&\textbf{95.8}& \textbf{95.7}     & \textbf{95.9}& 94.9&  \textbf{95.3}&53.2\\
        \bottomrule
    \end{tabular}
\end{table}

To address the competing needs of capturing global context and selectively filtering out noise, the AFIA module employs a novel dual-branch adaptive self-attention mechanism. This mechanism comprises two distinct branches: a Dense Self-Attention (DSA) branch and a ReLU-based Sparse Self-Attention (SSA) branch. We introduce ReLU-based sparse self-attention (SSA) to filter out negatively impacted features by low query-key matching scores. Concurrently, to prevent over-sparsity of ReLU-based self-attention, we introduce another dense self-attention branch (DSA), which employs the softmax layer, to aid in retaining crucial information. The key issue with this scheme is how to effectively reduce noisy and redundant information while retaining informative features as far as possible. Consequently, we employ an adaptive weight for each branch to facilitate their fusion. Furthermore, insufficient diversity in the input information or a lack of inter-channel interaction can weaken the learned feature representations. To address this, we incorporate channel shuffle, enhancing information exchange without significantly increasing computational cost. In conclusion, AFIA enhances feature representation and discriminative performance through its adaptive feature interaction and augmentation mechanisms.

In the upper branch of AFIA, which showed in Figure \ref{frame}, to further enhance feature diversity and prevent the AFIA module from relying too heavily on individual feature channels, we incorporate a channel shuffle operation. Traditional convolutional networks tend to learn redundant features across different channels. By shuffling feature channels, we encourage the model to learn more diverse and complementary features. Given a normalized feature map $X\in \mathbb{R}^{H \times W \times C}$, we first use a $1\times1$ convolution to adjust the channel dimensions. After this, a channel shuffling operation partitions the input
tensor along the channel axis into distinct groups. The resultant output tensors from each group are rearranged along the channel axis which engender a novel output tensor. Following this, we employ a stack of three $3\times3$ convolutional layers for feature extraction. The upper branch can be formulated as:
\begin{equation}
\begin{aligned}
    F_{cs} &= SConv_{3\times3}({Conv}_{1\times1}({CS}(X)), \\
\end{aligned}
\end{equation}
where $F_{cs}$ is an output of the upper branch, we define ${SConv}_{3\times3}$ as a $3\times3$ stacked convolution and ${Conv}_{1\times1}$ as a $1\times1$ convolution. ${CS(\cdot)}$ represents channel shuffle operation.

In the DSA branch, we use $3\times3$ depthwise convolutions after a $1\times1$ convolution to generate matrices of $queries$ $\mathbf{\mathit{Q}}$, $keys$ $\mathbf{\mathit{K}}$ and $values$ $\mathbf{\mathit{V}}$ from $X$. First, we revisit the standard dense self-attention mechanism (DSA) which is adopted in most existing works. It uses softmax over all query-key pairs for attention scores which can be formulated as:
\begin{equation}
\begin{aligned}
    DSA &= \text{Softmax}(\mathit{Q}\mathit{K}^{T}/\sqrt{d}+B), \\
\end{aligned}
\end{equation}
here $DSA$ denotes the dense self-attention mechanism; $\text{Softmax}(\cdot)$  is a softmax scoring function. $B$ refers to the learnable relative positional bias. Since not all query tokens are closely related to their corresponding key tokens, the SSA branch replaces the softmax function with a ReLU function, setting the attention scores of less relevant query-key pairs to zero. This significantly increases the sparsity of the attention map, effectively filtering out noise and interference from irrelevant features. This design is particularly crucial in colonoscopy video detection, selectively suppressing spurious activations caused by motion blur and specular reflections, and enabling the model to focus on the most discriminative features. The ReLU-based operation creates a sparse attention map, allowing the model to focus on the most informative features while suppressing noise. The formula is shown below:
\begin{equation}
\begin{aligned}
    SSA &= \text{ReLU}(\mathit{Q}\mathit{K}^{T}/\sqrt{d}+B), \\
\end{aligned}
\end{equation}
where $SSA$ denotes the sparse self-attention; $\text{ReLU}(\cdot)$ is the ReLU scoring function. 

While softmax-based DSA may introduce noisy interference from irrelevant regions, ReLU-based SSA can result in oversparsity, meaning that some learned features may not contain sufficient information for further processing. Thus, balancing these two mechanisms is an important issue. Therefore, we employ an adaptive feature fusion mechanism to effectively combine the features extracted by the DSA and SSA. This mechanism adaptively weights the outputs of each branch based on their relevance to the current input feature map. The weights for each branch are learned during training, allowing the model to dynamically adjust the importance of each branch based on the specific characteristics of the input data. It is worth noting that the trainable parameter is initialized with the outputs of each branch to enhance the adaptive rate. The formula for the weights and attention matrix can be formulated as:
\begin{equation}
\begin{aligned}
    w_n = \frac{e^{a_n}}{\sum_{i=1}^{N} e^{a_i}}, \quad n \in \{1, 2\}, \\
\end{aligned}
\end{equation}

\begin{equation}
\begin{aligned}
   F_{att} &= {Conv}_{1\times1}((w_1*DSA+w_2*SSA)\mathit{V}), \\
\end{aligned}
\end{equation}
here $\{a_i\}_{i=1,2}$ are learnable parameters initialized with the output of their corresponding branch. The $\{w_n\}_{n=1,2}$ are two normalized weights for adaptively modulating dual-branch. $F_{att}$ is the output of attention branches and $*$ denotes element-wise multiplication. Our AFIA achieves enhanced performance by establishing a superior trade-off between the suppression of noisy interactions from irrelevant areas and the leveraging of sufficient informative features.
    Finally, the output of the AFIA calculation is
computed as:
\begin{equation}
\begin{aligned}
    F_{out} &= F_{cs}+F_{att}. \\
\end{aligned}
\end{equation}

\subsubsection*{Scale-Aware Context Integration (SACI) Module} The feed-forward network (FFN) in traditional ViTs consists of linear layers for single-scale feature aggregation. However, the information contained in this single-scale feature aggregation is limited~\cite{r17}. Therefore, we leverage the SACI module to enhance non-linear feature transformation and aggregate multi-scale features.

Details of the SACI module are illustrated in Figure \ref{frame}. Input features are first dimension-adjusted using separate $1\times1$ convolutions before being processed by a dual-branch paradigm. In the upper branch, we use two $3\times3$ dilated convolutional layers with dilation rates $N1$ and $N2$ to facilitate the extraction of wider-ranging multi-scale features. In the lower branch, a stack of three 3×3 convolutional layers is used for feature extraction. Following this, nonlinear properties are introduced via the ReLU activation function. Given an input tensor $X\in \mathbb{R}^{H \times W \times C}$, SACI is formulated
as:

\begin{equation}
\begin{aligned}
F_{u} &= DConv^{N1}_{3\times3}({Conv}_{1\times1}(X))+DConv^{N2}_{3\times3}({Conv}_{1\times1}(X)),\\
\end{aligned}
\end{equation}
\begin{equation}
\begin{aligned}
F_{l} &= Conv_{1\times1}(SConv_{3\times3}({Conv}_{1\times1}(X))),
\end{aligned}
\end{equation}
\begin{equation}
\begin{aligned}
F_{s} &= F_{u}*\text{ReLU}(F_{l}).
\end{aligned}
\end{equation}
Where $F_{u}$ and $F_{l}$ means the output of the upper and lower branches. $DConv^{N1}_{3\times3}$ and $DConv^{N2}_{3\times3}$ denotes $3\times3$ dilated convolution
with dilation rate of $N1$ and $N2$. In the experimental setup, we assigned the values of 2 and 3 to $N1$ and $N2$. It is worth noting that the dilation rate should be determined based on the specific task requirements. Finally, $F_{s}$ is the output of our SACI moudel.

\begin{table}[t]
    \centering
    \caption{Quantitative comparison of different polyp detection approaches with our AVPDN on LDPolypVideo dataset and CVC-VideoClinicDB database.}
    \label{tab2}
    \begin{tabular}{l|cccc|cccc|c}
        \toprule
        \multirow{2}{*}{ Polyp Detection Approaches } & \multicolumn{4}{c|}{LDPolypVideo} & \multicolumn{4}{c|}{CVC-VideoClinicDB} &\multicolumn{1}{c}{}\\
        \cmidrule(lr){2-5} \cmidrule(lr){6-10}
                             & AP   & Precision & Recall & F1-Score  & AP   & Precision & Recall & F1-Score & FPS\\

        \midrule
 Colon SEG~\cite{r22}   & 67.3 & 67.5      & 65.3   & 66.3 & 77.7 & 78.9      & 88.4   & 83.4    &44.1\\
 STFT~\cite{r23}     & 65.6 & 74.5      & 50.6   & 60.2  & 90.8 & 92.6      & 92.3   & 92.4   &8.5\\
 ECC-PolypDet~\cite{r33}&68.5 &77.1 &53.9 &63.4 &91.0 &93.3 &92.8 & 93.0&46.1\\
 YONA~\cite{r11}& -& 75.4& 53.1& 62.3&- & 92.8& 93.8& 93.3&46.3\\
 AIDPT~\cite{r34}& -& -& -& -&- & 94.6& 84.5& 87.5&-\\ \midrule
        AVPDN (Ours)   & \textbf{96.6} & \textbf{96.8}& \textbf{95.0}&\textbf{95.8}& \textbf{95.7}     & \textbf{95.9}& 94.9&  \textbf{95.3}&53.2\\
        \bottomrule
    \end{tabular}
\end{table}

\section*{Experiment and Analysis}\label{sec:exp}  

To validate the proposed AVPDN superiority, it is compared with different image/video-based detectors and multiple state-of-the-art polyp detection approaches on the LDPolypVideo dataset~\cite{r36} and CVC-VideoClinicDB~\cite{r37} database. To ensure experimental fairness, we maintain consistent dataset settings for AVPDN and all other methods, following a 7:2:1 ratio for dividing the training, test, and validation sets.

\subsection*{Datasets}

\subsubsection*{LDPolypVideo dataset}
It comprises a comprehensive collection of colorectal polyp detection data, encompassing 160 endoscopic videos and 40,266 image frames. Critically, 33,884 of these frames contain at least one polyp, and the dataset annotates a total of 200 polyps. The dataset further includes 103 videos (comprising 861,400 frames) without complete polyp annotations, each labeled for the presence or absence of polyps. This unlabeled portion significantly enhances data diversity and allows for the application of unsupervised and semi-supervised learning approaches.

\subsubsection*{CVC-ClinicVideoDB database}
It is a publicly accessible database for advancing colorectal polyp detection research, comprising 40 endoscopic videos collected at the Hospital Clinic of Barcelona. It contains 17,000 frames, with 6,949 frames meticulously annotated for the presence of polyps. The dataset’s value lies in its comprehensive representation of polyp morphologies and sizes, alongside a diverse range of endoscopic scenarios, including variations in illumination, camera angles, and endoscope movement.

\subsection*{Evaluation Metrics}  
For the polyp detection task, we employ a suite of evaluation metrics to assess our model's performance rigorously. These metrics encompass several key components: precision, recall, average precision (AP), F1 score, and frames per second (FPS). Each metric provides a distinct perspective on the model’s ability to identify polyps while accurately minimizing false positives and negatives.

\subsubsection*{Precision} It represents the ratio of the number of samples correctly labeled as positive by the classifier to the number of all samples labeled as positive by the classifier. The formula is given by: 
\begin{equation}
Precision = \frac{TP}{TP + FP}.
\end{equation}
Where True Positives (\textit{TP }for abbreviation) are the correctly predicted positive instances, and False Positives (\textit{FP}) are the incorrectly predicted positive instances.

\subsubsection*{Recall} It represents the ratio of the number of samples correctly labeled as positive by the classifier to the number of all true positive samples. The formula is given by:
\begin{equation}
Recall = \frac{TP}{TP + FN}.
\end{equation}
False Negatives (FN) are the incorrectly predicted negative instances. 

\subsubsection*{Average Precision (AP)} For object detection tasks, AP is the area under the Precision-Recall curve, which is used to measure the average Precision under different Recalls. The formula is given by:
\begin{equation}
AP = \sum_{n} (P(n) \cdot \Delta \text{Recall}(n)),
\end{equation}
where P(n) is the Precision at the nth Recall value and \(\Delta\) Recall(n) is the difference between the nth Recall value and the n-1th Recall value. Typically, a Precision-Recall curve is obtained by sorting the model's output and calculating Precision and Recall at each threshold.

\subsubsection*{F1-score}
The harmonic mean of precision and recall, calculated as follows:
\begin{equation}
F_1 = 2 \cdot \frac{Precision \cdot Recall}{Precision + Recall}
\end{equation}

\subsubsection*{Frames Per Second (FPS)} FPS is a common evaluation metric used in the context of real-time video processing, object detection, and computer vision tasks, including the evaluation of polyp detection algorithms in colonoscopy videos.

\subsection*{Implementation Details} 
All computational experiments presented in this paper were conducted on a Linux server equipped with the PyTorch 2.11 framework, powered by 15 vCPU Intel(R) Xeon(R) Platinum 8474C processors and an NVIDIA GeForce RTX 4090D 24GB GPU. During training, all datasets were normalized and resized to a standard input image size of 640×640. In a series of preliminary experiments, we observed that the model converged within 60 epochs, with diminishing returns on further training. Therefore, we trained for 60 epochs on each
dataset. A batch size of 32 was employed for 60 epochs. An early stopping mechanism was implemented to mitigate overfitting, terminating training if the performance metric failed to demonstrate improvement for 100 consecutive epochs. The optimizer choice was AdamW. Optimal model performance was observed with a learning rate of 0.001 and a weight decay of 0.0005, as determined through a comprehensive grid search.

\subsection*{Comparison with State-of-the-art Methods} 
A quantitative comparison with state-of-the-art methods on the LDPolypVideo dataset and CVC-ClinicVideoDB database is presented in Table \ref{tab1} and Table \ref{tab2}. We can see that AVPDN demonstrates a clear performance advantage, outperforming all listed pure object detection methods on the LDPolypVideo dataset. Notably, when compared to the baseline model RT-DETR, our method achieves comprehensive improvements across all evaluation metrics. Specifically, it exhibits a gain of 2.4\% in Average Precision (AP) and precision, a 2.7\% improvement in recall, and a 2.5\% enhancement in the F1 score. These overall improvements can be mainly attributed to the design of our two novel modules, which highlight their effectiveness. 

The performance advantage becomes even more apparent when considering the results on the CVC-ClinicVideoDB database. Here, our approach achieves a 2.6\% increase in AP, reaching a high of 95.7\%, compared to both the second-best performing model YOLOv12 (AP: 93.1\%) and RT-DETR (AP: 93.1\%). These findings are particularly significant because the CVC-ClinicVideoDB is known for its diverse range of endoscopic scenarios, including variations in illumination, camera angles, and endoscope movement. This suggests that AVPDN is more robust and generalizable than existing approaches. It is worth noting that while previous methods for video polyp detection lacked precise polyp detection capabilities, AVPDN achieves the highest AP, precision, and F1 scores on the same challenging datasets.

Furthermore, when considering the comparison with specialized polyp detection methods like Colon SEG~\cite{r22}, STFT~\cite{r23}, ECC-PolypDet~\cite{r33}, YONA~\cite{r11} and AIDPT~\cite{r34} (Table \ref{tab2}), it is evident that AVPDN demonstrates superior performance in terms of the trade-off between precision and recall. Specifically, AVPDN achieves not only a high AP score but also maintains a high F1-score, indicating a balanced performance in minimizing both false positives and false negatives. This is critical in a clinical setting, where both types of errors can have significant consequences. For instance, a high false positive rate can lead to unnecessary biopsies, while a high false negative rate can result in missed cancerous or precancerous lesions. This robust performance indicates that the proposed method is superior to many state-of-the-art approaches in terms of detection accuracy, and it’s worth noting the table is also indicating the framerate (FPS) for each method to emphasize its real-time detection.

\begin{table}[t]
    \centering
    \caption{Ablation analysis of our baseline gradually including the newly proposed module and its components on LDPolypVideo dataset. AFIA means adaptive feature interaction and augmentation module; SACI means scale-aware context integration module; CS denotes channel shuffle; DSA and SSA stand for dense self-attention and sparse self-attention. }
    \label{tab3}
    \begin{tabular}{cccc|cccc}
        \toprule
         \multicolumn{3}{c}{AFIA} &\multicolumn{1}{c|}{SACI}& \multicolumn{4}{c}{LDPolypVideo} \\
         \cmidrule(lr){1-3} \cmidrule(lr){5-8}
         CS   & DSA & SSA & & AP   & Precision & Recall & F1-Score \\

        \midrule
         &       &       &  & 94.2 & 94.4& 92.3& 93.3\\ \midrule
         \checkmark &       &       &  & 94.6 & 94.6& 92.7   & $93.6_{\uparrow0.3}$\\
          & \checkmark&       &  & 95.5 & 95.9& 94.0& $94.9_{\uparrow1.6}$\\
          &       & \checkmark      &  & 95.2 & 96.0& 93.6& $94.7_{\uparrow1.4}$\\
          & \checkmark   & \checkmark      &  & 96.3 & 96.4& 94.7& $95.5_{\uparrow2.2}$\\
          \checkmark &\checkmark &\checkmark &  & 96.6 & 96.4& 94.9& $95.6_{\uparrow2.3}$\\
          &       &       &\checkmark  & 95.0 & 95.7& 93.8& $94.7_{\uparrow1.4}$\\
          \checkmark &\checkmark &\checkmark &\checkmark  & 96.6& 96.8& 95.0& $95.8_{\uparrow2.5}$\\
        \bottomrule
    \end{tabular}
\end{table}

\subsection*{Ablation Studies and Analysis}  
As demonstrated in Table \ref{tab3}, we conducted a series of ablation experiments to evaluate the individual contributions of the AFIA and SACI modules, which constitute the core innovations of the AVPDN framework. The results clearly show that both modules, along with their constituent components, are essential for optimizing the model’s overall polyp detection performance. This section provides a detailed analysis to elucidate the underlying mechanisms driving AVPDN’s superior performance.

\subsubsection*{Role of AFIA} The ablation studies presented in Table \ref{tab3} reveal that each component of the AFIA module contributes to improving the performance of the model. The integration of our proposed dual-branch adaptive self-attention mechanism (incorporating both DSA and SSA) into the baseline model yielded a significant performance boost, increasing the AP and F1-score by 2.1\% and 2.2\%, respectively. These results underscore the effectiveness of a synergistic DSA and SSA combination for enhanced feature extraction. The DSA branch, utilizing a softmax layer, helps to retain crucial contextual information, while the SSA branch, leveraging ReLU, effectively filters out noisy information from irrelevant regions. The fact that combining them adaptively yields a more significant improvement highlights the importance of balancing global context with noise suppression.

Furthermore, the introduction of a channel shuffle operation, designed to facilitate cross-channel information exchange, resulted in a substantial performance gain (F1-Score gain of 0.3\%) beyond that achieved with DSA and SSA alone. This improvement indicates that promoting interaction between different feature channels improves the diversity of feature representations, allowing the model to capture more complex and nuanced patterns in the data. Consequently, the complete AFIA module, integrating all three components (DSA, SSA, and Channel Shuffle), attained an overall F1-score of 95.6\%, a substantial improvement over the baseline. Taken together, these experimental findings demonstrate AFIA’s efficacy in mitigating background noise, promoting superior feature representations, and adaptively optimizing feature extraction for robust polyp detection.

\subsubsection*{Influence of SACI} The addition of the SACI module led to a marked increase in both AP and F1-score. Specifically, AP increased by 0.8\% and the F1-score increased by 1.4\%. This improvement underscores the influence of the SACI module in enhancing the polyp detection capabilities of our framework. The observed performance boost can be attributed to the SACI module’s ability to effectively integrate contextual information at multiple scales. By employing dilated convolutions with varying dilation rates, the SACI module enables the network to capture a broader receptive field, allowing it to understand the relationships between objects and their surrounding environment more effectively. For instance, the dilated convolutions enable the SACI module to capture long-range dependencies between different intestinal folds and vascular textures, which are important for distinguishing polyps from background structures. The substantial performance gain observed with the inclusion of the SACI module highlights its significant contribution to robust and accurate polyp detection by incorporating multi-scale contextual information. The performance is more robust on larger polyps with SACI, and on smaller polyps the AFIA module works in conjunction with the SACI module.

\begin{figure}[t]
    \centering
    \includegraphics[width=1\textwidth,height=0.7\textwidth]{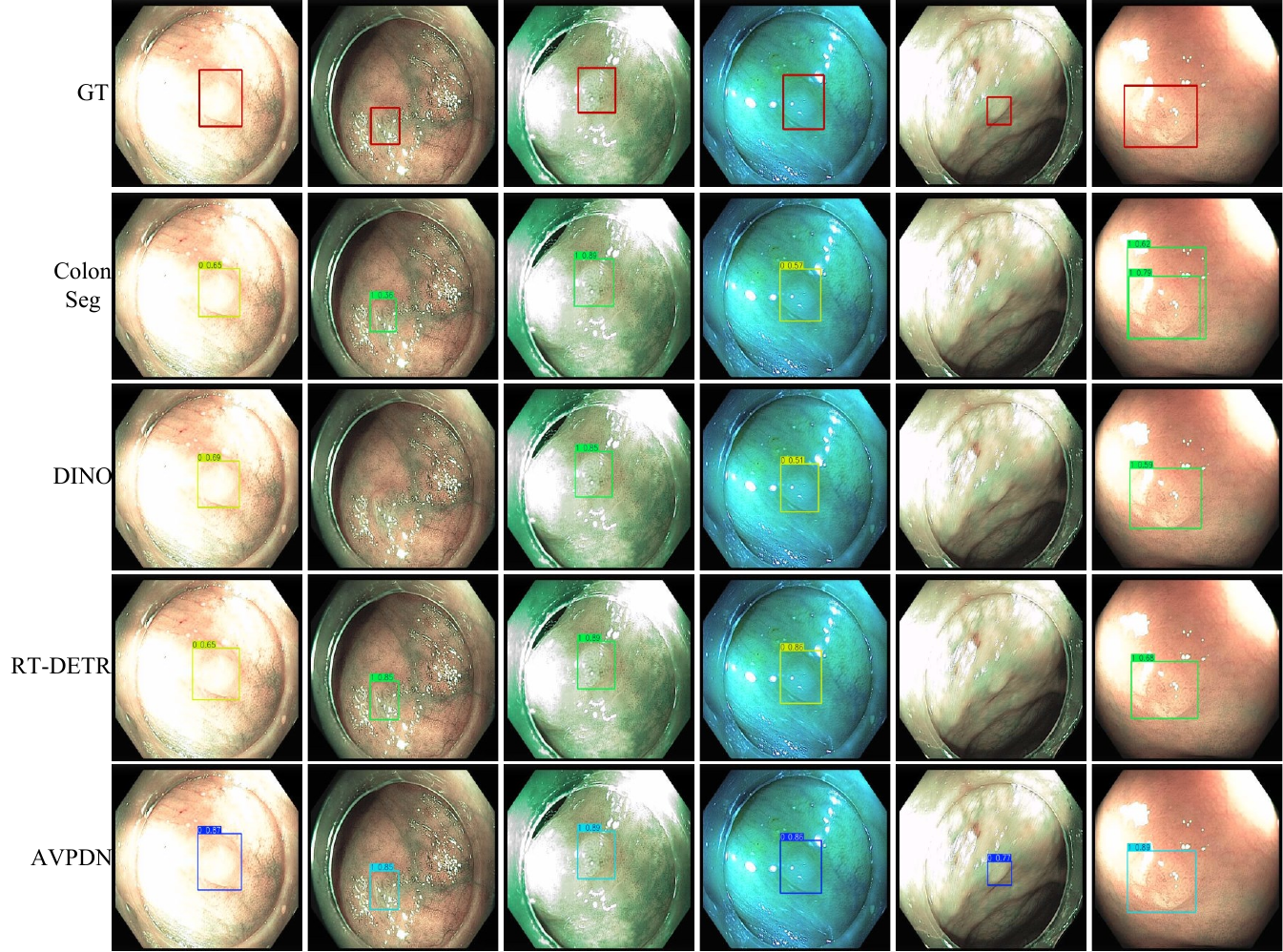}
\caption{Visualize results of baseline and AVPDN.The red box denotes ground truth. '$0$', the yellow box, and the dark blue box represent hyperplastic polyps, while '$1$', the green box, and the turquoise box represent adenomatous polyps.}
\label{vis}
\end{figure}
\subsection*{Visualization}  
The qualitative comparison, visually depicted in Figure \ref{vis}, reveals that AVPDN achieves superior performance relative to RT-DETR and other evaluated methods in challenging scenarios. The visualization highlights AVPDN’s ability to consistently generate accurate and well-defined bounding boxes, underscoring its effectiveness in detecting polyps despite diverse background disturbances, motion artifacts, and variations in polyp morphology and size. Noticeably, while baseline methods such as RT-DETR often struggle to accurately localize polyps, producing bounding boxes that are either poorly aligned with the polyp’s boundaries or fail to detect them altogether, AVPDN demonstrates a more robust and reliable detection capability. This improved performance can be attributed to AVPDN’s ability to focus on multi-scale polyp features, even in the presence of noise and complex intestinal environments. In particular, the enhanced feature representation achieved through the AFIA module enables the model to better distinguish polyps from surrounding tissues, while the SACI module facilitates the accurate detection of polyps across a range of scales. The improved accuracy in size, shape, and position of the bounding boxes generated by AVPDN ultimately leads to a more precise alignment with the ground truth annotations, demonstrating the practical advantages of our proposed framework for assisting clinicians in real-time polyp detection during colonoscopies. Specifically, we observe that in cases where polyps exhibit subtle visual differences from the surrounding mucosa, AVPDN is able to leverage the contextual information learned through its adaptive modules to achieve a more confident and accurate detection. Furthermore, AVPDN demonstrates a greater ability to distinguish between different types of polyps (e.g., hyperplastic vs. adenomatous), as indicated by the more accurate classification labels associated with the detected bounding boxes, which has significant implications for guiding downstream clinical decision-making.

\section*{Conclusion}
Dynamic colonoscopy videos provide more comprehensive polyp morphological information than static images, but their rapid motion characteristics challenge existing detection methods. This work addresses the challenge of accurate and efficient polyp detection in dynamic colonoscopy videos by proposing AVPDN, a novel framework characterized by the synergistic combination of AFIA and SACI modules.Specifically, AFIA effectively models global contextual dependencies and mitigates interference from less informative features, achieving adaptive optimization of feature representations.The SACI captures multi-scale spatial distribution characteristics of polyps by fusing coarse-grained global semantics and fine-grained local details, resulting in adaptive feature representation optimization. Experimental results across multiple public datasets demonstrate AVPDN’s superior performance in detection accuracy and related metrics compared to existing state-of-the-art techniques. AVPDN has demonstrated promising results, future work will focus on refining the architecture to improve computational efficiency and exploring its generalizability to other types of medical videos, such as those acquired during bronchoscopy or upper endoscopy. Additionally, investigating the interpretability of the learned features could provide valuable insights into the underlying mechanisms of polyp development and inform the design of even more effective detection strategies.

\section*{Data availibility}
The datasets used and analysed during the current study are available from the corresponding author on reasonable request.

\bibliography{sample}

\section*{Acknowledgements (not compulsory)}

This paper is the research result funded by Xi'an Shiyou University 2025 Graduate Innovation Fund Project (grant number YCX2513160). The funding amount is RMB 5,000.

\section*{Author contributions statement}

Zilin Chen completed methodology, software, writing-original draft, editing, visualization, project management, and resources. Shengnan Lu was responsible for supervision. All authors reviewed the manuscript.

\section*{Additional information}
The authors declare that they have no known competing financial interests or personal relationships that could have appeared to influence the work reported in this paper.

\end{document}